\documentclass[10pt,twocolumn,letterpaper]{article}

\usepackage[pagenumbers]{cvpr} 

\usepackage[dvipsnames, table]{xcolor}
\usepackage{graphicx}
\usepackage{multirow}
\usepackage{tabularx}
\usepackage{wrapfig}
\usepackage[acronym,toc]{glossaries}
\usepackage[flushleft]{threeparttable}

\definecolor{cvprblue}{rgb}{0.21,0.49,0.74}
\usepackage[pagebackref,breaklinks,colorlinks,citecolor=cvprblue]{hyperref}
\usepackage[most]{tcolorbox}
\usepackage{float}

\def\confName{CVPR}
\def\confYear{2025}

\title{Team RAS in 10th ABAW Competition:\\
Multimodal Valence and Arousal Estimation Approach
}

\author{Elena Ryumina\\
St. Petersburg Federal Research Center \\ of the Russian Academy of Sciences\\
St. Petersburg, Russia\\
{\tt\small ryumina.e@iias.spb.su}
\and
Maxim Markitantov\\
St. Petersburg Federal Research Center \\ of the Russian Academy of Sciences\\
St. Petersburg, Russia\\
{\tt\small markitantov.m@iias.spb.su}
\and
Alexandr Axyonov\\
St. Petersburg Federal Research Center \\ of the Russian Academy of Sciences\\
St. Petersburg, Russia\\
{\tt\small axyonov.a@iias.spb.su}
\and
Dmitry Ryumin\\
St. Petersburg Federal Research Center \\ of the Russian Academy of Sciences\\
St. Petersburg, Russia\\
{\tt\small ryumin.d@iias.spb.su}
\and
Mikhail Dolgushin\\
St. Petersburg Federal Research Center \\ of the Russian Academy of Sciences\\
St. Petersburg, Russia\\
{\tt\small dolgushin.m@iias.spb.su}
\and
Denis Dresvyanskiy\\
ITMO University\\
St. Petersburg, Russia\\
{\tt\small denisdresvyanskiy@gmail.com}
\and
Alexey Karpov\\
St. Petersburg Federal Research Center \\ of the Russian Academy of Sciences;\\
ITMO University \\
St. Petersburg, Russia\\
{\tt\small karpov@iias.spb.su}
}

\begin{document}
\newacronym{CER}{CER}{Compound Expression Recognition}
\newacronym{CE}{CE}{Compound Expressions}
\newacronym{ABAW}{ABAW}{Affective Behavior Analysis in-the-Wild}
\newacronym{AFEW}{AFEW}{Acted Facial Expressions in The Wild}

\newacronym{VS}{VS}{Static visual model}
\newacronym{VD}{VD}{Dynamic visual model}
\newacronym{FCL}{FCL}{Fully Connected Layer}
\newacronym{FPS}{FPS}{Frame Per Second}
\newacronym{LSTM}{LSTM}{Long Short-Term Memory}
\newacronym{VAD}{VAD}{Voice Activity Detection}
\newacronym{NLL}{NLL}{Negative Log-Likelihood}
\newacronym{SOTA}{SOTA}{State-of-the-Art}

\newacronym{Ne}{Ne}{Neutral}
\newacronym{An}{An}{Anger}
\newacronym{Di}{Di}{Disgust}
\newacronym{Fe}{Fe}{Fear}
\newacronym{Ha}{Ha}{Happiness}
\newacronym{Sa}{Sa}{Sadness}
\newacronym{Su}{Su}{Surprise}
\newacronym{ASR}{ASR}{Automatic Speech Recognition}

\newacronym{RAF-DB}{RAF-DB}{Real-world Affective Faces Database}
\newacronym{CLIP}{CLIP}{Contrastive Language-Image Pretraining}
\newacronym{Jina}{Jina}{Jina Embeddings V3}
\newacronym{RoBERTa}{RoBERTa}{Emotion English DistilRoBERTa Base}
\newacronym{MHPF}{MHPF}{Multi-Head Probability Fusion}
\newacronym{PF}{PF}{Probability Fusion}
\newacronym{LLM}{LLM}{Large Language Model}

\newacronym{PPA}{PPA}{Pair-Wise Probability Aggregation}
\newacronym{PFSA}{PFSA}{Pair-Wise Feature Similarity Aggregation}
\newacronym{MoE}{MoE}{Mixture-of-Experts}
\newacronym{ER}{ER}{Emotion Recognition}
\newacronym{VA}{VA}{valence-arousal}
\newacronym{GELU}{GELU}{Gaussian Error Linear Unit}
\newacronym{ReLU}{ReLU}{Rectified Linear Unit}
\newacronym{CCC}{CCC}{Concordance Correlation Coefficient}
\newacronym{E2E}{E2E}{End-to-End}
\newacronym{MAE}{MAE}{Mean Absolute Error}

\newacronym{ResNet}{ResNet}{Residual Network}
\newacronym{ViT}{ViT}{Visual Transformer}
\newacronym{TCN}{TCN}{Temporal Convolutional Network}
\newacronym{AE}{AE}{Auto-Encoder}

\newacronym{RJCA}{RJCA}{Recursive Joint Cross-Attention}
\newacronym{GR-JCA}{GR-JCA}{Gated Recursive Joint Cross Attention}
\newacronym{TAGF}{TAGF}{Time-aware Gated Fusion}
\newacronym{BERT}{BERT}{Bidirectional Encoder Representations from Transformers}
\newacronym{BiLSTM}{BiLSTM}{Bidirectional Long Short-Term Memory}
\newacronym{ITW}{ITW}{in-the-wild}
\newacronym{KELM}{KELM}{Kernel Extreme Learning Machine}
\newacronym{VLM}{VLM}{Visual Language Model}
\newacronym{PDEM}{PDEM}{Public Dimensional Emotion Model}

\newacronym{DCMMOE}{DCMMOE}{Directed Cross-Modal MoE} 
\newacronym{RAAV}{RAAV}{Reliability-Aware Audio-Visual}

\maketitle
\glsresetall
\begin{abstract}
Continuous emotion recognition in terms of valence and arousal under \gls{ITW} conditions remains a challenging problem due to large variations in appearance, head pose, illumination, occlusions, and subject-specific patterns of affective expression. We present a multimodal method for valence-arousal estimation \gls{ITW}. Our method combines three complementary modalities: face, behavior, and audio. The face modality relies on GRADA-based frame-level embeddings and Transformer-based temporal regression. We use Qwen3-VL-4B-Instruct to extract behavior-relevant information from video segments, while Mamba is used to model temporal dynamics across segments. The audio modality relies on WavLM-Large with attention-statistics pooling and includes a cross-modal filtering stage to reduce the influence of unreliable or non-speech segments. To fuse modalities, we explore two fusion strategies: a Directed Cross-Modal Mixture-of-Experts Fusion Strategy that learns interactions between modalities with adaptive weighting, and a Reliability-Aware Audio-Visual Fusion Strategy that combines visual features at the frame-level while using audio as complementary context. The results are reported on the Aff-Wild2 dataset following the 10th \gls{ABAW} challenge protocol. Experiments demonstrate that the proposed multimodal fusion strategy achieves a \gls{CCC} of 0.658 on the Aff-Wild2 development set.
\end{abstract}

\glsresetall

\section{Introduction}
\label{sec:intro}
This paper presents our contribution to the \gls{VA} task of the 10th edition of the \gls{ABAW} challenge series \cite{kollias2025dvd, kollias2024behaviour4all, kollias20247th, kollias20246th, kollias2024distribution, kollias2023abaw2, kollias2023multi, kollias2023abaw, kollias2022abaw, kollias2021analysing, kollias2021affect, kollias2021distribution, kollias2020analysing, kollias2019expression, kollias2019deep, kollias2019face, zafeiriou2017aff,  kollias2025emotions}. 
To replicate the results of our work, the reader is kindly referred to the GitHub repository\footnote{https://github.com/SMIL-SPCRAS/CVPRW-26}.

\gls{ER} is an important problem in Artificial Intelligence (AI) \cite{kim2025moss}. One promising direction is to model emotions in the continuous \gls{VA} space \cite{feldman1995valence}, where valence reflects pleasantness, and arousal indicates intensity of an emotion.

Significant progress in \gls{VA} estimation under \gls{ITW} conditions has been achieved in the \gls{ABAW} challenges, largely through deep neural and transformer-based multimodal models \cite{vaswani2017attention, kollias2025advancements}. However, the use of multimodal \gls{VLM} to derive behavior-oriented representations for \gls{VA} prediction remains underexplored, despite their promise of capturing contextual and situational affective cues \cite{ryumina2025zero, qiu2024language}.

In this work, we address continuous \gls{VA} estimation using three complementary sources of information: facial dynamics, acoustic cues, and behavior-oriented multimodal embeddings extracted with a \gls{VLM}. To combine them, we investigate two fusion strategies: a Directed Cross-Modal \gls{MoE}, where cross-modal Transformer experts model directed interactions between modality pairs and a gating mechanism adaptively weights them over time, and a Reliability-Aware Audio-Visual model, which performs frame-centric fusion of facial and behaviour features while using audio as auxiliary contextual evidence. These methods enable effective fusion of audio, face, and behavior-description representations for \gls{VA} prediction \gls{ITW} conditions.
\section{Related Work}
\label{sec:rw}
We compare our approach with State-of-the-Art (SOTA) methods proposed for the \gls{VA} task in the \gls{ABAW} competitions, including prior work from our team \cite{dresvyanskiy2024multi}.

Among unimodal methods, video-based approaches have shown consistently strong performance. The baseline system \cite{kollias20246th} used cropped and aligned 112×112 face images with a 50-layer \gls{ResNet}. Later work explored lightweight visual architectures such as Mobile \gls{ViT} \cite{mehta2021mobilevit}, MobileFaceNet \cite{chen2018mobilefacenets}, EfficientNet \cite{tan2019efficientnet}, and DDAMFN \cite{zhang2023dual}, together with temporal smoothing to reduce frame-level noise \cite{savchenko2023emotieffnets, savchenko2025leveraging}. The strongest unimodal solutions in the 8th \gls{ABAW} challenge was reported by \citet{zhou2025emotion}, who combined a Masked \gls{AE} with \gls{CLIP} and a \gls{TCN} for temporal modeling. 

Multimodal video-audio methods have also been extensively studied. \citet{dresvyanskiy2024multi} explored intermediate cross-attention and late fusion, combining transformer-based \gls{PDEM} audio features\footnote{https://github.com/audeering/w2v2-how-to} with visual backbones and a functional-based \gls{KELM}. Likewise, \citet{zhang2024effective} combined a Masked \gls{AE} \cite{he2022masked} for vision and VGGish \cite{chen2020vggsound} for audio with an ensemble of transformer encoders, achieving \gls{SOTA} performance in the 6th \gls{ABAW} \gls{VA} task. 

Another important direction has focused on modality interaction with cross-attention and gating. \citet{rajasekhar2025united} proposed \gls{GR-JCA} in the 8th \gls{ABAW} challenge, introducing adaptive and hierarchical gating for audio-visual fusion. In the same competition, \citet{yu2025interactive} presented the winning method, combining \gls{ResNet}-based visual features with VGGish and LogMel audio representations, followed by \gls{TCN}-based temporal modeling and cross-modal attention. This line was further extended by \citet{lee2025dynamic} in the 9th \gls{ABAW} challenge through \gls{TAGF}, which integrated \gls{TCN}s, \gls{RJCA}, and \gls{BiLSTM}-based temporal gating to better handle dynamic emotional changes and cross-modal misalignment.

Overall, despite extensive research on attention, gating and temporal modeling the reviewed studies did not employ \gls{VLM}s for continuous \gls{VA} estimation. In contrast, our approach uses a multimodal \gls{VLM} to extract behavior segment-level embeddings and integrates them with face and audio representations within a cross-modal fusion strategy.

\section{Proposed Method}
\label{sec:method}

\begin{figure*}
  \centering
   \includegraphics[width=0.95\linewidth]{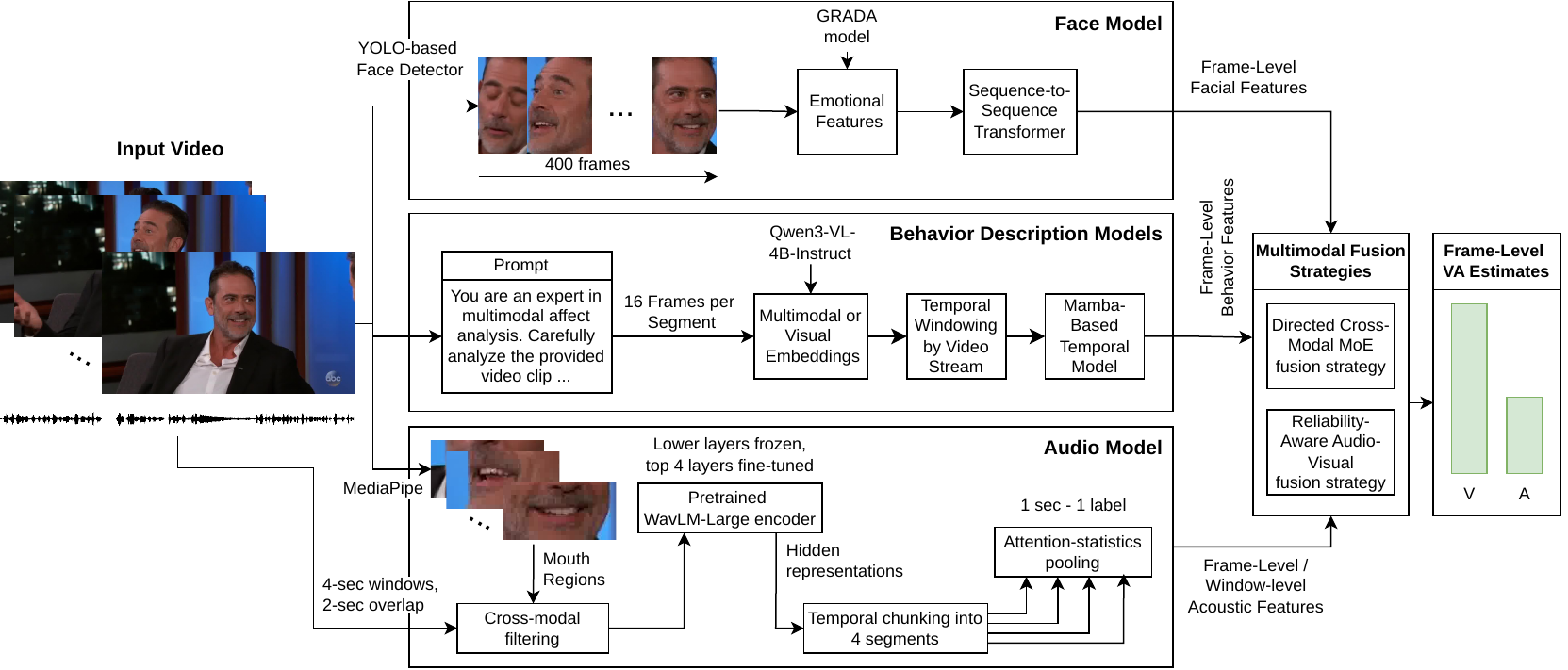}
   \caption{Pipeline of the proposed multimodal method for \gls{VA} estimation.}
   \label{fig:pipeline}
\end{figure*}

The pipeline of the proposed multimodal method for \gls{VA} estimation is shown in Figure~\ref{fig:pipeline}.

\subsection{Face Model}

We first detect face regions in each video frame using a YOLO-based face detector\footnote{https://github.com/lindevs/yolov8-face}. Because some clips contain multiple visible faces and identity switches across frames, we apply manual identity curation to remove mismatched or ambiguous tracks. This step is necessary to ensure that the final face stream corresponds to a single target subject per video sequence. For each curated face crop, we extract frame-level affective representations using the GRADA encoder~\cite{dresvyanskiy2024multi} described below. The preprocessing follows the GRADA training protocol, including: (i) resizing the images while preserving the aspect ratio; (ii) zero-padding the images to a fixed square resolution of $240 \times 240$ pixels; (iii) applying ImageNet~\cite{deng2009imagenet} normalization to the images.

The GRADA model is based on the EfficientNet-B1~\cite{tan2019efficientnet} architecture, pre-trained on ImageNet~\cite{deng2009imagenet} and afterward fine-tuned as a multi-task model for facial expression recognition on a large-scale mixed dataset combining ten publicly available affective corpora (the details on the training procedure can be found in~\cite{dresvyanskiy2024multi}). During the fine-tuning, the original classification head was replaced with a 256-dimensional embedding layer followed by two branches for emotion classification and regression tasks (valence and arousal).

The fine-tuning employed discriminative learning rates and gradual blockwise unfreezing, along with extensive data augmentation to improve robustness under \gls{ITW} conditions. Among many neural network architectures evaluated, EfficientNet-B1 demonstrated the best trade-off between generalization and computational efficiency while remaining compact (7.9M parameters). In this work, the 256-dimensional embedding layer serves as the frame-level affective feature extractor.

Temporal modeling is performed using a transformer-based sequence regression model. Given frame-level feature vectors, the model uses: (i) a projection block that consists of a \gls{FCL}, layer normalization, and dropout; (ii) multiple transformer layers~\cite{vaswani2017attention}; and (iii) a regression head that includes a \gls{FCL}, layer normalization, \gls{GELU} activation, dropout, and final \gls{FCL}. 

Each video is converted into overlapping temporal windows of fixed length $L$ and stride $S$. If a video tail is not exactly divisible, an additional final window is added to ensure coverage of the last frames. This preserves temporal continuity while increasing the number of training samples. If a requested frame feature is unavailable (face not found), the face model pipeline uses nearest-frame. 

\subsection{Behavior Description Models}

We use the Qwen3-VL-4B-Instruct\footnote{https://huggingface.co/Qwen/Qwen3-VL-4B-Instruct} model (hereafter referred to as Qwen3) as a multimodal feature encoder to capture behavior-related cues. For each video segment, the model processes 16 uniformly thinned frames together with an affect-oriented textual prompt, guiding the model to focus on behavior-relevant information, including facial expressions, head movements, gestures, posture, and scene context. The full prompt is shown in Figure~\ref{fig:affect_prompt}.


\begin{figure}
\centering
\begin{tcolorbox}[
    title=Prompt for behavior description,
    colback=white,
    colframe=gray!60,
    colbacktitle=gray!30,
    coltitle=black,
    fonttitle=\bfseries,
    boxrule=0.4pt,
    arc=0pt,
    left=6pt,
    right=6pt,
    top=6pt,
    bottom=6pt,
    width=\columnwidth
]
\small
You are an expert in multimodal affect analysis. Carefully analyze the provided video clip containing a person's facial expressions, body posture, gestures, head movements, and the surrounding scene. Describe the person's affect in terms of valence (pleasant $\leftrightarrow$ unpleasant) and arousal (low $\leftrightarrow$ high). You may mention uncertainty or mixed or ambiguous affect when appropriate. Explain which visible cues support your valence and arousal estimates and how they evolve over time. Avoid assumptions beyond what is visually present. Write a single coherent paragraph with no line breaks, no bullets, no special formatting, no longer than 100 tokens, and always end with a period.
\end{tcolorbox}
\caption{Prompt used for behavior description extraction with Qwen3.}
\label{fig:affect_prompt}
\end{figure}

With hidden state extraction enabled, the final behavior embedding is denoted by $e \in \mathbb{R}^{d}$. In our configuration, $e$ is extracted from the last hidden layer using the last valid token representation, yielding a fixed-dimensional representation for each video segment. We consider two extraction settings, namely visual and multimodal. In the visual setting, the embedding is derived from visual tokens only, whereas in the multimodal setting it is derived from the joint video-text input. These segment-level embeddings are then used as inputs to the downstream temporal model for continuous \gls{VA} estimation.

To model temporal dynamics across segments, the extracted embeddings are grouped by video stream and organized into temporal sequences, which can additionally be split into fixed-length windows with a predefined stride. The resulting sequences are projected into a shared latent space and processed by a stack of Mamba blocks, which combine local temporal filtering with state-space sequence modeling to capture both short-term affective fluctuations and longer-range dependencies. The final hidden representations are then mapped to valence and arousal estimates through a regression head. 

During frame-level evaluation, segment-level predictions are expanded to the frame interval covered by each segment. If several overlapping segments contribute to the same frame, their predictions are averaged, yielding a continuous frame-level \gls{VA} trajectory.

\subsection{Audio Model}
For audio, we propose a continuous~\gls{VA} estimation model. First, each video is converted into 4-sec segments with 2-sec overlap, mono, 16 kHz, and padded or truncated to a fixed length. Before training, we apply a sample filtering stage based on visual mouth-opening dynamics~\cite{dresvyanskiy2024multi} extracted from the corresponding video using MediaPipe~\cite{lugaresi2019mediapipe}. Short gaps between open-mouth regions are temporally smoothed, isolated short open-mouth bursts are suppressed, and an audio segment is retained only if the accumulated open-mouth duration and annotation coverage exceed predefined thresholds. This cross-modal filtering is designed to approximate speech presence in a video-oriented Aff-Wild2~\cite{kollias2019expression} dataset, where audio is often noisy and unreliable. This criterion is not fully reliable, but it helps filter out many segments that are unlikely to contain actual speech.

The acoustic encoder is based on WavLM-Large~\cite{chen2022wavlm} model, which is a self-supervised speech model. We use pretrained model\footnote{https://huggingface.co/3loi/SER-Odyssey-Baseline-WavLM-Multi-Attributes} trained on MSP-Podcast~\cite{lotfian2019building} for the Odyssey 2024 Emotion Recognition competition~\cite{goncalves24_odyssey}, which estimates arousal, valence, and dominance. To adapt the model to the target task while reducing overfitting, we fine-tune only the top four transformer layers of the WavLM backbone, while keeping the remaining layers frozen.

The sequence of hidden representations produced by WavLM is divided into four equal temporal chunks to match the sequence-to-sequence target format (4 seconds - 4 labels). Each chunk is aggregated with attention-statistics pooling, where trainable attention weights are used to compute a weighted mean and a weighted standard deviation of the frame-level features. The pooled chunk descriptors are then passed through a shared regression head composed of normalization, dropout, and linear projection layers. The final \gls{FCL} produces two outputs corresponding to valence and arousal.

\subsection{Modality Fusion Strategies}
\textbf{\gls{DCMMOE} Fusion Strategy.} We propose the \gls{DCMMOE} Fusion Strategy for continuous \gls{VA} estimation. Each modality is first projected into a shared latent space, after which all ordered pairs of modalities form cross-attention experts (one modality serves as query, the other as key/value), explicitly modeling asymmetric inter-modal interactions. A learnable gating network then assigns adaptive weights to the expert outputs for each segment to allow the model to prioritize the most informative cross-modal relations under varying signal quality, and the fused representation is decoded into final \gls{VA} estimation.

Let $\mathcal{M}=\{1,\dots,M\}$ denote the set of modalities.
For each modality $m\in\mathcal{M}$, the input is a frame-level sequence
\[
\mathbf{X}^{(m)}\in\mathbb{R}^{B\times L\times d_m},
\]
where $B$ is batch size, $L$ is temporal length, and $d_m$ is modality-specific feature dimensionality. Each modality is first mapped to a shared latent dimension $d_h$:
\[
\mathbf{H}^{(m)} = \phi_m(\mathbf{X}^{(m)}),\quad
\phi_m:\mathbb{R}^{d_m}\to\mathbb{R}^{d_h},
\]
where $\phi_m$ is a projection block that consists of a \gls{FCL}, layer normalization, and dropout.

We define an expert for each ordered pair of distinct modalities:
\[
\mathcal{E}=\{(q,k)\mid q\in\mathcal{M},\,k\in\mathcal{M},\,q\neq k\},
\]
where the total number of experts is $|\mathcal{E}|=M(M-1)$ experts.
Expert $f_{(q,k)}$ performs stacked cross-attention with
query from modality $q$ and key/value from modality $k$:
\[
\mathbf{Z}_{(q,k)} = f_{(q,k)}\!\left(\mathbf{H}^{(q)},\mathbf{H}^{(k)},\mathbf{H}^{(k)}\right).
\]
Each expert $f_{(q,k)}$ is implemented as a stack of $N$ cross-attention transformer layers with $H$ attention heads.
For each time step $l$, the gating logits are calculated from the mean projected multimodal state:
\[
\bar{\mathbf{h}}_l=\frac{1}{M}\sum_{m\in\mathcal{M}}\mathbf{H}^{(m)}_l,\qquad
\mathbf{g}_l=\mathbf{W}_g\bar{\mathbf{h}}_l+\mathbf{b}_g \in \mathbb{R}^{|\mathcal{E}|}.
\]
The fused representation is the expert-weighted sum:
\[
\mathbf{z}_{\text{fused},l} = \sum_{(q,k)\in\mathcal{E}}
softmax(\mathbf{g}_{(q,k),l})\,\mathbf{Z}_{(q,k),l},
\]
\[
\mathbf{z}_{\text{fused},l}\in\mathbb{R}^{d_h}.
\]

Finally, a linear regression head performs \gls{VA} estimation.

This design explicitly models directed cross-modal interactions (query modality and context modality) and learns data-dependent expert weighting per frame via gating, enabling adaptive fusion under modality-specific uncertainty.

\textbf{\gls{RAAV} Fusion Strategy.} We propose the \gls{RAAV} Fusion Strategy, a frame-centric multimodal architecture for \gls{VA} estimation in sliding windows. Facial and behavior multimodal features are fused at each frame by masked reliability-aware gating, while audio is used as an auxiliary context through a small set of learnable bottleneck latent representations. For frame $l$, the visual fusion is defined as follows:

\[
\mathbf{h}_l^{(m)} \in \mathbb{R}^{d_h},
\qquad
\alpha_l^{(m)} \in \mathbb{R},
\]
\[
\alpha_l^{(m)}=\operatorname{MaskedSoftmax}\!\left(g_m(\mathbf{h}_l^{(m)})+\log \pi_m\right),
\]
\[
\mathbf{z}_{\mathrm{vis},l}=\sum_{m}\alpha_l^{(m)}\mathbf{h}_l^{(m)},
\qquad
\mathbf{z}_{\mathrm{vis},l}\in\mathbb{R}^{d_h},
\]

where $\mathbf{h}_l^{(m)}$ is the modality representation, $g_m(\cdot)$ is a learned scoring function, $\pi_m$ is a modality prior, and invalid modalities are excluded by masking, $\mathbf{z}_{\mathrm{vis},l}$ is the fused visual token.

The fused visual sequence then attends to the audio bottleneck representation ($\mathbf{B}_{a}$):
\[
\mathbf{Z}_{0}=\operatorname{LN}\!\Big(
\mathbf{Z}_{\mathrm{vis}}+
\operatorname{CrossAttn}(\mathbf{Z}_{\mathrm{vis}},\mathbf{B}_{a},\mathbf{B}_{a})
\Big),
\]
where LN is layer normalization.

A lightweight transformer encoder and regression head finally performs frame-level \gls{VA} estimation. This asymmetric design reflects the nature of the task: visual modalities determine the temporal resolution, while audio provides complementary window-level evidence.

\section{Experiments}

\begin{table*}[t]
\centering
\resizebox{\textwidth}{!}{
\begin{tabular}{@{}llrrrr@{}}
\toprule
\multirow{2}{*}{ID} & \multirow{2}{*}{Model} & \multicolumn{3}{c}{Aff-Wild2 (devel. set)} & \multicolumn{1}{c}{Aff-Wild2 (test set)} \\
\cmidrule(l){3-5} \cmidrule(l){6-6}
 & & Valence & Arousal & Avg. & Avg. \\
\midrule
1 & Video-based model. GRADA + Transformer & 0.5869 & 0.6508 & 0.6189 & 0.54\\
\multirow{2}{*}{2} & Visual-Based Behavior Description Model. & \multirow{2}{*}{0.2499} & \multirow{2}{*}{0.5515} & \multirow{2}{*}{0.4007} & \multirow{2}{*}{--} \\
 & Qwen3 + Mamba & & & & \\
\multirow{2}{*}{3} & Multimodal-Based Behavior Description Model. & \multirow{2}{*}{0.4290} & \multirow{2}{*}{0.6480} & \multirow{2}{*}{0.5385} & \multirow{2}{*}{--} \\
& Qwen3 + Mamba & & & & \\
4 & Audio-based model. WavLM + Chunk-Wise Pooling & 0.3415 & 0.4636 & 0.4025 & -- \\
\bottomrule
5&Model IDs 1 \& 4 (frame-level) \& \gls{DCMMOE}&\textbf{0.6252}& 0.6671& 0.6461 & 0.58  \\
6&Model IDs 1 \& 2 \& 4 (frame-level) \& \gls{DCMMOE}&0.5647& 0.6797& 0.6222 & --  \\
7&Model IDs 1 \& 3 \& 4 (frame-level) \& \gls{DCMMOE}&0.6100& 0.6875 & 0.6487 & 0.61  \\
8 & Model IDs 1 \& 3 \& 4 (window-level) \& \gls{RAAV}& 0.6078 & \textbf{0.7073} & \textbf{0.6576} & 0.62 \\
\bottomrule
\end{tabular}}
\caption{Experimental results on Aff-Wild2 \gls{VA} estimation.}
\label{tab:Results}
\end{table*}



\subsection{Research Data}
For all experiments, we used the Aff-Wild2 dataset with the official valence-arousal annotations and subject-independent splits provided in the 10th ABAW Challenge. Aff-Wild2 is an audiovisual in-the-wild dataset characterized by large variations in subject appearance, head pose, illumination, occlusions, and recording conditions, making continuous \gls{VA} estimation particularly challenging. For the \gls{VA} task, the dataset contains 594 videos, including 16 videos with two annotated subjects, resulting in approximately 3 million annotated frames from 584 subjects. Valence and arousal are provided as continuous frame-level labels in the range [-1,1], and the official split includes 356 training, 76 development, and 162 test videos.

\subsection{Experimental Setup}
The official evaluation measure for the \gls{VA} Estimation challenge is the \gls{CCC}. \gls{CCC} additionally penalizes discrepancies in the mean values of predictions and targets, making it more suitable for continuous affect estimation~\cite{lawrence1989concordance}. The measure is defined as follows:
\[
    CCC = \frac{2\cdot \sigma_{t,p}}{\sigma_{t}^2+\sigma_{p}^2+(\mu_{t}-\mu_{p})^2},
\]
where $\mu_{t}$ and $\mu_{p}$ are the means of the ground truth and predicted scores, respectively; $\sigma_{t}$ and $\sigma_{p}$ denote the respective standard deviations; $\sigma_{t,p}$ are their variances, and $\sigma_{t,p}$ denotes the covariance between $t$ and $p$.

Model optimization was performed using a hybrid loss based on \gls{CCC}, with an optional \gls{MAE} term used in some experiments to improve training stability. We additionally varied the relative weights of the loss components as well as the valence/arousal weighting scheme across experiments.

During the training process of the face model and the \gls{DCMMOE} fusion strategy, the optimal parameters $L=400$, $S=150$, $N=5$, $H=16$ were selected by grid search. The learning rate was $1e-4$, the batch size was 8, the AdamW optimizer with weight decay equal to 0.01, and the ReduceLROnPlateau learning rate schedule were set the same for all these models.

For the Mamba temporal model trained on Qwen3 visual embeddings, the best configuration was obtained by grid search using embeddings from the last hidden state. The embeddings were arranged into temporal windows of length 16 with stride 8, and processed by a 4-layer Mamba model with hidden dimensionality 128, state size 8, kernel size 3, and a regression head of size 512. Training used batch size 8, AdamW, a learning rate of $1e\text{-}4$, weight decay of $1e\text{-}4$, and dropout of 0.2 throughout the model.

For the Mamba temporal model trained on Qwen3 multimodal embeddings, we used the multimodal representation from the last hidden state. The embeddings were arranged into temporal windows of length 16 with stride 8, and processed by a 12-layer Mamba model with hidden dimensionality 256, state size 8, kernel size 5, and a regression head of size 512. Training used batch size 8, AdamW, a learning rate of $3e\text{-}4$, weight decay of $1e\text{-}3$, and dropout of 0.2 throughout the temporal model.

In audio-based \gls{VA} estimation, we explored different pretrained models, their configurations, training hyperparameters, optimization strategies, as well as loss configurations. Training was performed for 50 epochs with batch size 8, AdamW optimization, a backbone learning rate of $5e-6$, a head learning rate of $2e-4$, and weight decay of 0.01.

\subsection{Experimental Results}
The results in Table~\ref{tab:Results} show clear performance differences between unimodal and multimodal configurations for Aff-Wild2 \gls{VA} estimation. Among the unimodal models, the video model (Model ID 1) achieves the strongest overall result, with an average \gls{CCC} of 0.6189 on the development set, while the audio model and the visual Qwen3 model perform substantially worse. The multimodal Qwen3 model (Model ID 3) clearly outperforms its visual counterpart (Model ID 2), indicating that behavior multimodal embeddings provide richer affective information than visual representations. Across the unimodal models, arousal is estimated more reliably than valence, whereas the video model remains the most balanced.

Multimodal fusion consistently improves the results over the individual models, confirming the effectiveness of the proposed \gls{DCMMOE} Fusion Strategy. In particular, combining the video, multimodal Qwen3, and audio models (Model IDs 1, 3, and 4) yields the best frame-level fusion result, with an average \gls{CCC} of 0.6487 and the highest arousal score of 0.6875. The same trend is observed on the test set, where performance increases from 0.54 for the unimodal video model to 0.58 for video-audio fusion and to 0.61 for the three model fusion. The \gls{RAAV} Fusion Strategy further improves the validation results, achieving the highest arousal score of 0.7073 and the highest average \gls{CCC} of 0.6576, as well as an average \gls{CCC} of 0.62 on the test set. Overall, these results show that the proposed fusion strategies effectively exploit complementary information from face, audio, and behavior embeddings, while achieving performance competitive with existing \gls{SOTA} approaches for \gls{VA} estimation.

\section{Conclusions}
In this paper, we proposed a novel multimodal approach for continuous \gls{VA} estimation under \gls{ITW} conditions that combines three complementary modalities: visual, behavior multimodal embeddings and audio. The visual modality is modeled using GRADA-based frame-level embeddings and Transformer-based temporal regression, the behavior modality relies on Qwen3-VL-4B-Instruct with Mamba-based temporal modeling, and the audio modality is built on WavLM-Large with chunk-wise attention-statistics pooling and cross-modal filtering of unreliable segments. To integrate these modalities, two fusion strategies were investigated: a \gls{DCMMOE} Fusion Strategy, which models directed inter-modal interactions with adaptive gating, and the \gls{RAAV} Fusion Strategy, which performs frame-centric fusion of visual and behavior features while using audio as complementary contextual evidence.

Multimodal fusion consistently outperformed the unimodal models, while the best results were obtained with the \gls{RAAV} Fusion Strategy, highlighting the importance of adaptive multimodal integration for continuous affect estimation. The proposed method achieved an average \gls{CCC} of 0.6576 on the Aff-Wild2 development set and 0.62 on the test set. Furthermore, the Qwen-based multimodal behavior model clearly outperformed its visual counterpart, demonstrating the value of multimodal \gls{VLM} behavior embeddings for \gls{VA} estimation. In general, the obtained performance is competitive with the \gls{SOTA} approaches, which confirms the effectiveness of the proposed multimodal framework.
{
    \small
    \bibliographystyle{ieeenat_fullname}
    \bibliography{main}
}


\end{document}